\documentclass[a4paper,10pt]{article}
\usepackage[utf8]{inputenc}
 \usepackage[english]{babel}
\usepackage{graphicx}
\usepackage{a4wide}

\usepackage{authblk}
\usepackage{amssymb}
\usepackage{epsfig}
\usepackage{subfigure}
\usepackage{calc}
\usepackage{amstext}
\usepackage{amsmath}
\usepackage{amsthm}
\usepackage{multicol}
\usepackage{pslatex}
\usepackage{bbm}
\usepackage[small]{caption}
 \usepackage[english]{babel}
 \usepackage{amsfonts}
\usepackage{stmaryrd} 
\usepackage{xspace}
 \usepackage{mathtools}
 \usepackage{cite}
 
 \SetSymbolFont{stmry}{bold}{U}{stmry}{m}{n} 
 
\begin{document}
\title{Predicting Alzheimer's disease: a neuroimaging study with 3D convolutional neural networks}  

\author[1]{Adrien Payan} 
\author[1,2]{Giovanni Montana\footnote{Corresponding author: \tt{giovanni.montana@kcl.ac.uk}}}
\affil[1]{Department of Mathematics, Imperial College London, London SW7 2AZ, UK}
\affil[2]{Department of Biomedical Engineering, King's College London, St Thomas' Hospital, London SE1 7EH, UK}


\maketitle

\begin{abstract}

Pattern recognition methods using neuroimaging data for the diagnosis of Alzheimer's disease have been the subject of extensive research in recent years. In this paper, we use deep learning methods, and in particular sparse autoencoders and 3D convolutional neural networks, to build an algorithm that can predict the disease status of a patient, based on an MRI scan of the brain. We report on experiments using the ADNI data set involving 2,265 historical scans. We demonstrate that 3D convolutional neural networks outperform several other classifiers reported in the literature and produce state-of-art results.

\end{abstract}

\section{\uppercase{Introduction}}\label{sec:introduction}

\let\thefootnote\relax\footnote{*Data used in preparation of this article were obtained from the Alzheimer's Disease Neuroimaging Initiative (ADNI) database (adni.loni.usc.edu). As such, the investigators within the ADNI contributed to the design and implementation of ADNI and/or provided data but did not participate in analysis or writing of this report. A complete listing of ADNI investigators can be found at: http://adni.loni.usc.edu/wp-content/uploads/how\_to\_apply/\\ADNI\_Acknowledgement\_List.pdf}

\noindent Alzheimer's Disease (AD) is the most common type of dementia. Dementia refers to diseases that are characterized by a loss of memory or other cognitive impairments, and is caused by damage to nerve cells in the brain. In the United States, an estimated 5.2 million people of all ages have AD in 2014. Mild cognitive impairment (MCI) is a condition in which an individual has mild but noticeable changes in thinking abilities. Individuals with MCI are more likely to develop AD than inviduals without \cite{AD}.

Early detection of the disease can be achieved by magnetic resonance imaging (MRI), a technique that uses a magnetic field and radio waves to create a detailed 3D image of the brain. A multitude of machine learning methods have been tried for this task in recent years, including support vector machines, independent component analysis and penalized regression. Some of these methods have been shown to be very effective in diagnosing AD from neuroimages, sometimes even more effective than human radiologists. For instance, recent studies have shown that machine learning algorithms were able to predict AD more accurately than experienced clinicians \cite{kloppel}. It is therefore of great interest to
develop and improve such prediction methods.

In this paper we build a learning algorithm that, using MRI images as input, is able to discriminate between healthy brains (HC) and diseased brains. We investigate a class of deep artificial neural networks, and specifically a combination of sparse autoencoders and convolutional neural networks. The main novelty of our approach is to use 3D convolutions on the whole MRI image, which yield better performance than 2D convolutions on slices in our experiments. We report on classification results obtained using a 3-way classifier (HC vs. AD vs. MCI) and three binary classifiers (AD vs. HC, AD vs. MCI and MCI vs. HC).

This article is organised as follows. We describe the data in Section \ref{sec:data}, and introduce the deep learning approach in Section \ref{sec:methods}. Section \ref{sec:review} offers a brief review of different classification methods that have been reported in the literature for this problem. Finally, in Section \ref{sec:results}, we provide the experimental results and a discussion.

\section{\uppercase{Experimental Data}}\label{sec:data}

\begin{figure*}
\centering
\includegraphics[width = \textwidth]{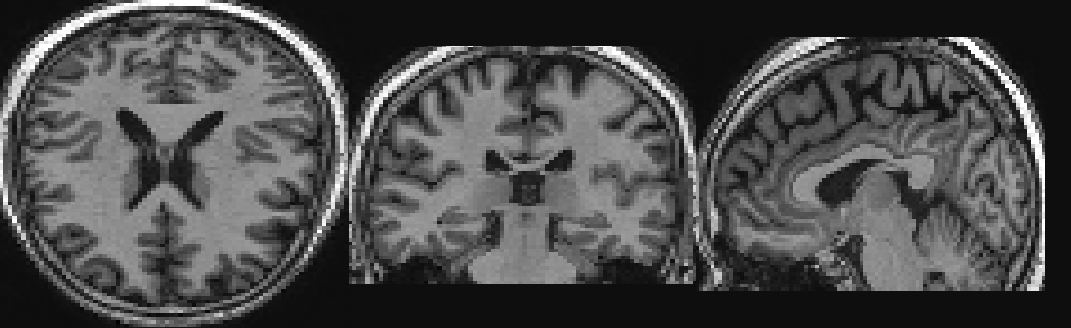}
\caption{Slices of an MRI scan of an AD patient, from left to right: in axial view, coronal view and sagittal view.}
\label{brain-slices}
\end{figure*}

\noindent For our experiments we use MRI data made available as part of the Alzheimer's Disease Neuroimaging Initiative (ADNI). ADNI is an ongoing, multicenter study designed to develop clinical, imaging, genetic, and biochemical biomarkers for the early detection and tracking of Alzheimer's disease. The ADNI study began in 2004 and is now in its third phase. The dataset used here was originally prepared and analysed in \cite{gupta} and consists of 755 patients in each one of the three classes (AD, MCI, HC), for a total of 2,265 scans. Statistical Parametric Mapping (SPM) was used to normalize the image data into an International Consortium for Brain Mapping template. The configuration includes a positron density template with no weighting image, and a 7th-order B-spline for interpolation whilst the remaining parameters were set to their default. We also normalised the data by subtracting the mean and dividing by the standard deviation. The dimension of each image is $68 \times 95 \times 79$, which results in 510,340 voxels. Figure \ref{brain-slices} shows an example of three two-dimensional slices extracted from an MRI scan.

\section{\uppercase{Deep Neural Networks}}
\label{sec:methods}

\noindent We take a two-stage approach whereby we initially use a sparse autoencoder to learn filters for convolution operations, and then build a convolutional neural network whose first layer uses the filters learned with the autoencoder. In this paper we are particularly interested in comparing the performance of 2D and 3D convolutional networks. In Section 3.1 we present the architecture of the sparse autoencoder, and in Section 3.2 we describe the 3D convolutional network. The 2D network is detailed in Section 3.2. 

\subsection{Sparse Autoencoder}

\noindent An autoencoder is a 3-layer neural network which is used to extract features from an input such as an image \cite{practical}. The autoencoder has an input layer, a hidden layer and an output layer; each layer contains several units. The input and output layers have the same number of units. This network has an encoder function $f$, which maps a given input $x \in \Re^{n}$  to a hidden representation $h \in \Re^{p}$, and a decoder function $g$, which maps the representation $h$ to the output $\hat{x} \in \Re^{n}$. In our problem the inputs $x$ are 3D patches extracted from the scans. The purpose of the decoder function is to reconstruct the input $x$ from the hidden representation $h$. Let $W \in \Re^{p\times n}$ and $b \in \Re^{p}$ be the matrix of weights and the vector of biases of the encoder function, respectively; analogously, let $W^{*} \in \Re^{n\times p}$ and $b^{*} \in \Re^{n}$ be the weights and biases of the decoder function, respectively. The autoencoder estimates the parameters in  
\begin{equation}
h = f(Wx+b)
\end{equation}
\begin{equation}
\hat{x} = g(W^{*}h+b^{*})
\end{equation}
where $f$ is the sigmoid function, and $g$ is the identity function. Here we use the identity function for the decoder because the inputs are real-valued, whereas a sigmoid function would constrain the output units to be in the interval $[0,1]$, and the reconstruction would be more difficult. Moreover, we impose tied weights $W^{*} = W^{T}$. For real-valued inputs such as pixel intensities, an appropriate choice for a cost function is the mean squared error, i.e.
\begin{equation}
J(W,b) = \frac{1}{N}\sum_{i=1}^{N}\frac{1}{2} \| \hat{x}^{(i)} - x^{(i)}\|^2
\end{equation}
where $N$ is the total number of inputs.
The autoencoder can be used to obtain a new representation of the input data through its hidden layer. We decided to try using an autoencoder with an overcomplete hidden layer, i.e. an autoencoder which has an equal or larger number of hidden units than input units. Autoencoders with overcomplete hidden layers can be useful feature extractors. One potential issue with overcomplete autoencoders is that if we only minimize the reconstruction error, then the hidden layer can potentially just learn the identity function \cite{practical}. We therefore need to impose additional constraints. In our experiments we use autoencoders with sparsity constraints \cite{sparse}. We investigate whether a sparse autoencoder, i.e. an autoencoder obtained by enforcing most hidden units to be close to zero, can be used to learn useful filters for convolution operations. The sparsity constraint is expected to be advantageous in this context because it encourages representations that may disentangle the underlying factors controlling the variability of MRI images.

Let $a_{j}(x)$ denote the activation of hidden unit $j$ when the autoencoder is given input $x$, and let
\begin{equation}
\hat{s}_{j} = \frac{1}{N}\sum_{i=1}^{N}\left[a_{j}(x^{(i)})\right]
\end{equation}
be the mean activation of hidden unit $j$, averaged over the training set. We try to impose the constraint $\hat{s}_{j} = s$ for all $j$, where $s$ is a sparsity hyper-parameter, typically a small value close to zero (e.g. $s = 0.05$). To satisfy this constraint, we add a penalty term to our cost function. We choose a penalty based on the concept of Kullback-Leibler divergence:
\begin{equation}
\sum_{j=1}^{h}KL(s||\hat{s}_{j})
\end{equation}
where
\begin{equation}
KL(s||\hat{s}_{j}) = s\log(\frac{s}{\hat{s}_{j}}) + (1-s)\log(\frac{1-s}{1-\hat{s}_{j}})
\end{equation}
quantifies the divergence between a Bernoulli distribution with mean $s$ and one with mean $\hat{s}_{j}$. In the sum, $h$ is the number of units in the hidden layer, and the index $j$ is summing over all the hidden units. This penalty has the property that $\text{KL}(s||\hat{s}_{j}) = 0$ if $\hat{s}_{j} = s$, and otherwise it increases as $\hat{s}_{j}$ moves away from $s$. Thus, minimizing this term has the effect of causing $\hat{s}_{j}$ to be close to $s$. The cost function we use here is
\begin{equation}
J_{2}(W,b) = J(W,b) + \beta\sum_{j=1}^{h}KL(s||\hat{s}_{j}) + \lambda\sum_{i,j}W_{i,j}^{2}
\end{equation}
where $J(W,b)$ is as defined previously, and $\beta$  is a hyper-parameter which controls the weight of the penalty term. We have also added a third term to the cost function, which is called weight decay, and is used to reduce overfitting; $\lambda$ is a hyper-parameter which controls the amount of weight decay.

In our approach, we train an autoencoder on a set of randomly selected 3D patches of size $5 \times 5 \times 5 = 125$ extracted from the MRI scans. The purpose of this autoencoder training is to learn filters for convolution operations: as we will explain in section 3.2, a convolution covers a series of spatially localised regions in an input. In total, we extract $1,000$ patches from $100$ scans in the training set, so we have a total of $100,000$ patches. We train a sparse overcomplete autoencoder with $150$ hidden units on this set of patches. We use 80,000 patches for the training set, 10,000 patches for the validation set and 10,000 patches for the test set. Each patch is unrolled into a vector of size 125.

The cost function is minimised using gradient descent with mini batches \cite{practical}: the training set is divided into several mini batches, and at each iteration we use only one of these mini batches in the function that we minimize; the algorithm is expected to converge faster than with full batches.

We also define as a basis the set of all the weights linking one unit in the hidden layer to all the units in the input layer. A basis will try to extract spatially localised features in the input. The bases are used in the next section on convolutional neural networks.

\subsection{3D Convolutional Networks}

\noindent After training the sparse autoencoder, we build a 3D convolutional network which takes as input an MRI scan. Convolutional networks have been found to be useful for image classification problems in several domains such as handwritten digit recognition \cite{jarrett} and object recognition \cite{imagenet}.

These artificial neural networks are made up of convolutional, pooling and fully-connected layers. The networks are characterised by three main properties: local connectivity of the hidden units, parameter sharing and the use of pooling operations.

We first describe the idea of local connectivity. In a hidden layer, a unit is not connected to all the units in the previous layer, but only to a small number of units in a spatially localised region. This property is beneficial in a number of ways. On one hand, it reduces the number of parameters, thus making the architecture less prone to overfitting whilst also alleviating memory and computational issues. On the other hand, by modelling portions of the image, the hidden units are able to detect local patterns and features which may be important for discrimination. The part of the image that a hidden unit is connected to is referred to as the "receptive field". Each possible receptive field of a fixed size is associated with a specific hidden unit. The set of all these hidden units corresponds to a single "feature map", that is a set of hidden units which altogether cover the whole image. 

A hidden layer has several feature maps and all the hidden units within a feature map share the same parameters. This parameter sharing feature is useful because it further reduces the number of parameters, and because hidden units within a feature map extract the same features at every position in the input.

Let $y^{k}$ be the 3-dimensional array of the $k^{th}$ feature map in a hidden layer, and $x$ be the 3-dimensional array of the input. Also let $W_{k}$ be the 3-dimensional filter connecting the input to the $k^{th}$ feature map, and let $b_{k}$ be the scalar bias term for the $k^{th}$ feature map. The computation of a feature map is given by:
\begin{equation}
y^{k} = f(W_{k}*x + b_{k})
\end{equation}
where $f$ is the sigmoid activation function, and $*$ denotes the convolution operation. In the computation the scalar term $b_{k}$ is added to every entry of the array $W_{k}*x$. We assume that $W_{k}$ is of size $r \times s \times t$. We define the convolution of an input $x$ with a filter $W_{k}$ as $[W_{k}*x](i,j,k)$, which is
\begin{equation}
\sum_{u=0}^{r-1}\sum_{v=0}^{s-1}\sum_{w=0}^{t-1}W_{k}(r-u,s-v,t-w) x(i+u,j+v,k+w)
\end{equation}
where $M(i,j,k)$ denotes the $(i,j,k)$ entry of a 3D array $M$. The convolution of an input map of size $\textit{m} \times \textit{p} \times \textit{q}$ with a filter of size $\textit{r} \times \textit{s} \times \textit{t}$ gives an output of size $\textit{(m-r+1)} \times \textit{(p-s+1)} \times \textit{(q-t+1)}$.

A layer which is made up of several feature maps obtained this way is called a convolutional layer.

For every basis of the sparse autoencoder that we trained previously, we use the set of learned weights of that basis as a 3D filter of a 3D convolutional layer. By applying the convolutions with all the bases, we obtain a convolutional layer of 150 3D feature maps. Since the patches are of size $5 \times 5 \times 5$, a convolution of an image with a basis produces a feature map of size $(68-5+1)\times(95-5+1)\times(79-5+1) = 64\times91\times75$. We also add the bias term associated with the basis and apply a sigmoid activation function to every unit in the feature map. This convolutional layer is likely to discover local patterns and structures in the 3D input image: it allows the algorithm to exploit the 3D topology/spatial information of the image.

Convolutional layers are followed by pooling layers. We use max-pooling, which consists in segmenting each feature map into several non-overlapping and adjacent neighbourhoods of hidden units. Within every neighbourhood, only the hidden unit with the largest activation (i.e. the maximum) is retained. The pooling operation reduces the number of units in a hidden layer, which is useful as noted above. Pooling also builds robustness to small distortions of the image such as translations. In our approach, we apply a $5\times5\times5$ max-pooling operation to reduce the size of the feature maps of the convolutional layer. Each feature map therefore becomes a max-pooled feature map of size $(64/5)\times(91/5)\times(75/5) = 12\times18\times15$, where we round down to the nearest integer because we ignore the borders.

\begin{figure*}
	\centering
	\includegraphics[width = \textwidth]{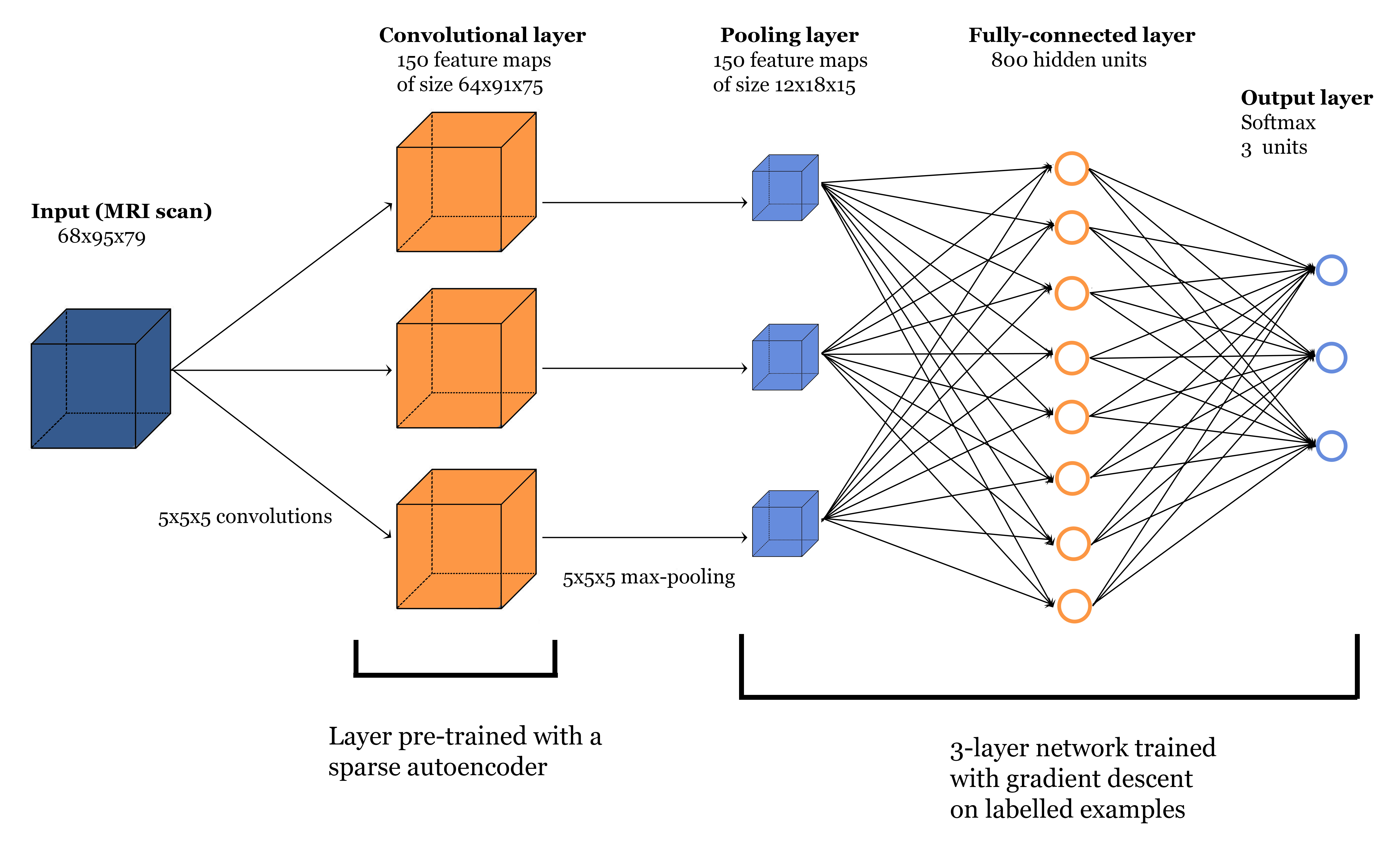}
	\caption{Architecture of the neural network used for 3-way classification. For binary classification, the output layer only has two units.}
\label{architecture}
\end{figure*}

The outputs of every max-pooled feature map are then stacked. With $150$ feature maps of size $12\times18\times15$, there is a total of $150\times12\times18\times15 = 486,000$ outputs. These outputs are used as inputs for a $3$-layer fully-connected neural network (i.e. with an input, hidden and output layer). We choose a hidden layer with $800$ units with a sigmoid activation function, and an output layer with 3 units with a softmax activation function. The 3 units in the output layer represent the conditional probabilities that the input belongs to each of the classes (AD, MCI, HC). Figure \ref{architecture} provides an illustration of the network architecture.

We take the cross-entropy as our cost function $J(W,b)$, which is commonly used for classification tasks \cite{glorot}. This function is:
\begin{equation}
-\frac{1}{N} \sum_{i=1}^{N} \sum_{j=1}^{3} \left[\mathbbm{1}\{y^{(i)} = j\} \log (h_{W,b}(x^{(i)})_{j})\right]
\end{equation}
where $N$ is the number of MRI scans, $j$ is summing over the 3 classes, $h_{W,b}$ is the function computed by the network and $x^{(i)}, y^{(i)}$ are the input and label of the $i^{th}$ scan, respectively. We do not use weight decay because in early experiments the addition of this term was not found to be beneficial.

The 3-layer network is trained with mini batch gradient descent. The weights of the hidden layer are randomly initialised, and the weights of the softmax layer are initialised to zero. It is important to note that we do not include the convolutional layer in the final training; the convolutional layer is only "pre-trained" with an autoencoder. We also use the momentum method to speed up the training of the 3-layer network. Briefly, this method consists in adding a weighted average of past gradients in the gradient descent updates so as to remove some noise, particularly in directions of high curvature of the cost function \cite{practical}. 

We use the validation set to determine the early-stopping time: at the end of the optimisation algorithm, we keep the network parameters which led to the lowest validation error during the course of the algorithm. Then we evaluate the performance of the network on the test set.

\subsection{2D Convolutions}

\noindent In our experiments we also test an approach using 2D convolutions for comparative purposes. Our initial hypothesis was that the 3D approach would provide a boost in performance compared to more traditional 2D convolutions as capturing local 3D patterns and structures in the image can be useful for discrimination. 

The 2D approach consists in training a sparse autoencoder on 2D patches extracted from slices of the MRI scans. In this case, we extract patches of size $11\times11=121$. We again use $150$ hidden units for the autoencoder, exactly as in the 3D approach. We then apply 2D convolutions on all $68$ slices of a scan in order to obtain a feature map: since each slice has size $95\times79$, a convolution of a slice with an autoencoder basis gives us an output of size $(95-11+1)\times(79-11+1)=85\times69$. As in the previous architecture, we also add a bias term and use a sigmoid activation function. The max-pooling operation in this case consists of  $10\times10$ patches, and reduces the size of a slice to $(85/10)\times(69/10)=8\times6$. The outputs of all the max-pooled feature maps are then stacked. Since we have $68$ slices for each feature map, we obtain a total of $150\times68\times8\times6 = 489,600$ outputs, which is comparable in size to the previous architecture. We deliberately choose to have similar sizes for the outputs of the 2D and 3D approaches in order to make the comparison of the two as accurate as possible. These outputs are then used as inputs in a 3-layer fully connected network. All other hyper-parameters are kept the same as before. 

\section{\uppercase{A brief review of MRI classifiers for AD}}
\label{sec:review}

\noindent Several pattern classifiers have been tried for the discrimination of subjects using structural MRI modalities in AD. Among the many approaches, support vector machines (SVM) have been used extensively in the area. Table \ref{review} contains a selection of related studies along with the sample sizes and the reported performance. Although a direct comparison of these studies is difficult, as each study uses different datasets and preprocessing protocols, the tables gives an indication of typical accuracy measures achieved in the classification of MRI images.

\cite{kloppel} used SVMs with linear kernels for the classification of grey matter signatures, and benchmarked their results against the performance achieved by expert radiologists, which surprisingly were found to be less accurate than the algorithm. Another approach used independent component analysis (ICA) as a feature extractor, coupled with a SVM algorithm \cite{ica}. \cite{montana} describe an approach that combines penalised regression and data resampling for feature extraction prior to the classification using SVMs with Gaussian kernels. \cite{batmanghelich} report that best performance is achieved using a SVM classifier with the bagging method (for AD vs HC) and a logistic regression model with a boosting algorithm (for MCI vs HC). \cite{instance} extract highly discriminative patches which are then classified using a SVM with graph kernels; using methods such as t-tests or sparse coding, each patch is assigned a probability which quantifies its discriminative ability.

Very recently, deep learning methods have also been explored for MRI data classification. In \cite{gupta}, an autoencoder is first used to learn features from 2D patches extracted from either MRI scans or natural images. The parameters of the autoencoder are then used as filters of a convolutional layer. As in our study, the classification is achieved using a 3-layer neural network with softmax function. The difference between their network architecture and ours is the use of 3D convolutions. \cite{sydney} report on a deep fully-connected network pre-trained with stacked autoencoders which is then fine-tuned; however, their approach does not use convolution operations, contrary to ours.

\begin{table*}
\centering
\caption{Review of selected methods for the AD and MCI classification. We provide the sample size and the accuracy of every method. The accuracy refers to the proportion of correct predictions on the test set. AD is Alzheimer's disease; MCI is mild cognitive impairment; HC is healthy control. For example, AD vs. HC means that we are trying to classify scans as AD or HC.}
\label{review}
\begin{tabular}{p{30mm} | p{50mm} | p{65mm}}
	\textbf{Paper} & \textbf{Sample sizes} & \textbf{Accuracy} \\ \hline
	\cite{gupta} & 755 from each class (AD, MCI, HC) & with natural image patches:\newline AD vs. HC 94.74\%\newline MCI vs. HC 86.35\%\newline AD vs. MCI 88.10\%\newline 3-way 85\% \newline with MRI patches:\newline AD vs. HC 93.80\% \newline MCI vs. HC 83.30\% \newline AD vs. MCI 86.30\% \newline 3-way 78.20\% \\ \hline
	\cite{sydney} & 65 AD, 67 cMCI, 102 ncMCI, 77 HC & AD vs. HC 87.76\%\newline MCI vs. HC 76.92\%\newline 4-way 47.42\% \\ \hline
	\cite{kloppel} & Several groups (first: 20 AD, 20 HC; second: 18 AD, 19 FTLD; third: 14 AD, 14 HC) & AD vs. HC: up to 95\% accuracy (range = 87.5-95\%) depending on the patient group\\ \hline
	\cite{montana} & Baseline MRI: 198 AD, 409 MCI (pMCI and sMCI), 231 HC\newline Longitudinal MRI: 510 subjects & Baseline MRI:\newline AD vs. HC 87.9\%\newline pMCI vs. HC 83.2\%\newline pMCI vs. sMCI 70.4\%\newline
Longit MRI:\newline AD vs. HC 90.3\%\newline pMCI vs. HC 86.9\%\newline pMCI vs. sMCI 82.1\% \\ \hline
	\cite{ica} & 202 AD, 410 MCI, 236 HC & 75\% of data in training set:\newline AD vs. HC 78.4\%\newline MCI vs. HC 71.2\%\newline 90\% of data in training set:\newline AD vs. HC 85.7\%\newline MCI vs. HC 79.2\% \\ \hline
	\cite{batmanghelich} & 56 AD, 60 MCI, 60 HC & AD vs. HC 89\% \newline MCI vs. HC 72\% \\ \hline
	\cite{instance} & 198 AD, 238 sMCI, 167 pMCI, 234 HC & AD vs. HC 88.8\% \newline sMCI vs. pMCI 69.6\%
\end{tabular}
\end{table*}

\section{\uppercase{Results}}
\label{sec:results}

\noindent The convolutional neural networks using 2D and 3D convolutions were trained used a training set of $1,731$ examples. A validation set of $306$ examples was used to determine the early-stopping time as explained above. Finally, a test set of $228$ examples was used to evaluate the performance of the model on unseen examples, and compute the performance figures reported below.

Table \ref{tab:results} gives the accuracy (i.e. the proportion of correct predictions) of the 2D and 3D architectures. As expected, the 3D approach has a superior performance for the 3-way comparison, as well as the AD vs. MCI and HC vs. MCI comparisons. For the AD vs. HC comparison, there are no noticeable differences. Our results compare very favourably with those in Table \ref{review} reported by other studies, although no definite claims can be made about the superiority of our approach due to differences in datasets, sample sizes and preprocessing steps. 

\begin{figure*}
	\centering
	\includegraphics[width = \textwidth]{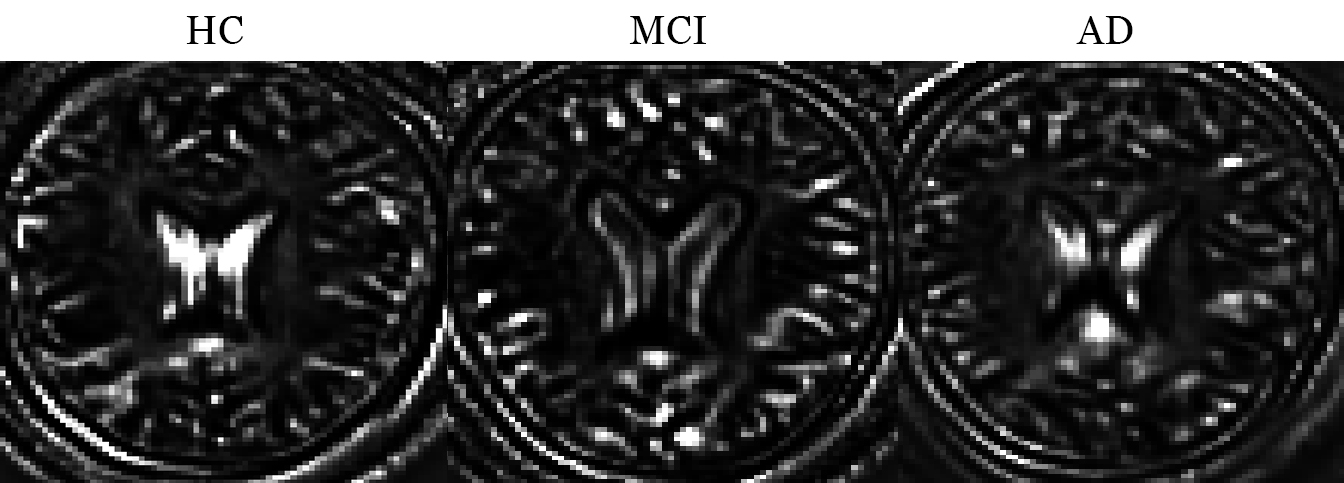}
	\caption{Examples of convolutions with the fourth basis of the 3D sparse autoencoder (32nd slice). An example from each class is randomly chosen.}
\label{convolutions}
\end{figure*}

The interpretation of our results is made difficult by the nature of the deep neural network architectures. Figure \ref{convolutions} shows the convolutions of MRI scans from each of the 3 classes with the fourth basis of the 3D sparse autoencoder.

\begin{table}[h]
\caption{Results of the models on the test set. The accuracy refers to the proportion of correct predictions. 3-way means that we are trying to classify a scan as AD, MCI or HC, and the other three lines (AD vs. HC, AD vs. MCI and HC vs. MCI), refer to binary classifications. The second column corresponds to results with 2D convolutions, and the third column corresponds to results with 3D convolutions.}
\label{tab:results} \centering
\begin{tabular}{l l l}
	\textbf{Classification} & \textbf{Accuracy (2D)} & \textbf{Accuracy (3D)} \\ \hline
	3-way & 85.53\% & 89.47\% \\
	AD vs. HC & 95.39\% & 95.39\% \\
	AD vs. MCI & 82.24\% & 86.84\% \\
	HC vs. MCI & 90.13\% & 92.11\% \\
	\hline
\end{tabular}
\end{table}

\section{\uppercase{Conclusion}}
\label{sec:conclusion}

\noindent In this paper we designed and tested a pattern classification system that combines sparse autoencoders and convolutional neural networks. We were primarily interested in assessing the accuracy of such an approach on a relatively large patient population, but we also wanted to compare the performance of 2D and 3D convolutions in a convolutional neural network architecture. Our experiments indicate that a 3D approach has the potential to capture local 3D patterns which may boost the classification performance, albeit only by a small margin. These investigations could be further improved in future studies by carrying out more exhaustive searches for the optimal hyper-parameters in both architectures. Moreover, the overall performance of these systems could be further improved. For instance, the convolutional layer used in our experiments has been pre-trained with an autoencoder, but not fine-tuned. There is evidence that fine-tuning may improve the performance \cite{jarrett} at the cost of a much increased computational complexity at the training stage.

\section*{\uppercase{Acknowledgements}}

\noindent We would like to thank Ashish Gupta for sharing the preprocessed ADNI data.

\bibliographystyle{plain}
{\small
\bibliography{DLAD}}

\end{document}